\pdfoutput=1

\documentclass[11pt]{article}

\usepackage[]{acl}

\usepackage{times}
\usepackage{latexsym}
\usepackage{amsmath}
\usepackage{tabularx}
\usepackage{color}

\usepackage[T1]{fontenc}

\usepackage[utf8]{inputenc}

\usepackage{microtype}

\usepackage{inconsolata}

\usepackage{graphicx}
\usepackage{multirow}

\title{PUMGPT: A Large Vision-Language Model for Product Understanding}

\author{
Wei Xue$^{1}$, Zongyi Guo$^{2}$, Baoliang Cui$^{2}$, Zheng Xing$^{2}$, Xiaoyi Zeng$^{2}$, Xiufei Wang$^{2}$, Shuhui Wu$^{1}$, Weiming Lu$^{1\dagger}$\\
$^{1}$College of Computer Science and Technology, Zhejiang University, China \\
$^{2}$Alibaba Group, China\\
\texttt{\{lokilanka, shwu, luwm\}@zju.edu.cn} \\
\texttt{\{zongyi.gzy, moqing.cbl, xingzheng.xz, xiufei.wxf\}@alibaba-inc.com} \\
\texttt{yuanhan@taobao.com} \\
}

\begin{document}
\maketitle
\let\thefootnote\relax

\begin{abstract}
E-commerce platforms benefit from accurate product understanding to enhance user experience and operational efficiency. Traditional methods often focus on isolated tasks such as attribute extraction or categorization, posing adaptability issues to evolving tasks and leading to usability challenges with noisy data from the internet. Current Large Vision Language Models (LVLMs) lack domain-specific fine-tuning, thus falling short in precision and instruction following. To address these issues, we introduce \textbf{\textsc{PumGPT}}, the first e-commerce specialized LVLM designed for multi-modal product understanding tasks. We collected and curated a dataset of over one million products from AliExpress, filtering out non-inferable attributes using a universal hallucination detection framework, resulting in 663k high-quality data samples. \textbf{\textsc{PumGPT}} focuses on five essential tasks aimed at enhancing workflows for e-commerce platforms and retailers. We also introduce \textbf{\textsc{PumBench}}, a benchmark to evaluate product understanding across LVLMs. Our experiments show that \textbf{\textsc{PumGPT}} outperforms five other open-source LVLMs and GPT-4V in product understanding tasks. We also conduct extensive analytical experiments to delve deeply into the superiority of \textsc{PumGPT}, demonstrating the necessity for a specialized model in the e-commerce domain.
\end{abstract}

\section{Introduction}
\begin{figure}[t]
  \includegraphics[width=\columnwidth]{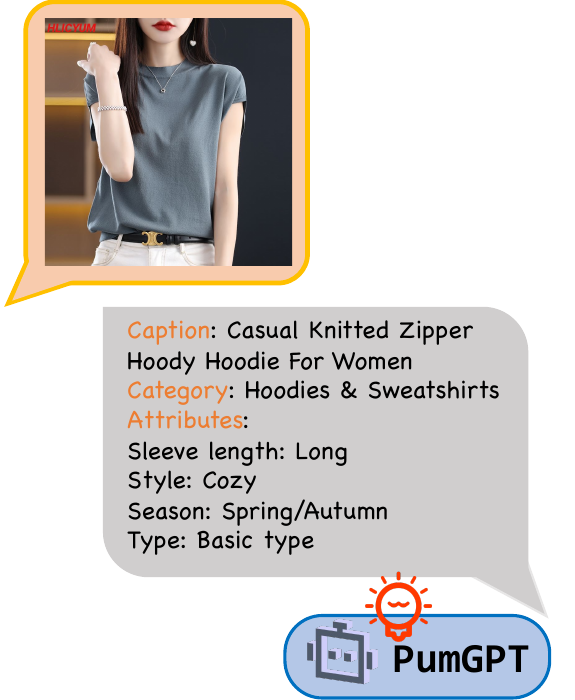}
  \caption{A glimpse on \textsc{PumGPT} in product understanding.}
  \label{fig:fig1.1}
\end{figure}

\footnote{$\dagger$Corresponding author}
E-commerce platforms extensively rely on a deep understanding of products to boost online shopping experiences. As is shown in Figure \ref{fig:fig1.1}, for instance, given a product image, the ability to automatically generate appealing caption, accurately categorize the product, and extract its attributes not only improves product recommendation\citep{le2021explainable, sun2020lara} and product search\citep{prodsearchahuja2020, ai2017learning} on platforms but also facilitates retailers to launch and update their goods with substantial time savings.

Nevertheless, traditional methods typically focus only on a subset of tasks within a series of product understanding tasks. For instance, they may solely address product attribute extraction\citep{shinzato2022simple, yan2021adatag, zou2024implicitave} or categorization tasks\citep{lin2021pam}. Training a specific model for each task proves challenging to adapt to ever-evolving tasks and new products and diminishes usability.
Moreover, the product attribute data scraped from the Internet contains a significant amount of noise\citep{wang2020learning, zhu2020multimodal, yang2022mave}. For example, certain attribute values cannot be inferred from the product captions and images since some retailers might supplement the attributes with information not present in the images or captions. Directly training models with such dirty samples can lead to severe hallucination problems\citep{zhu2023vdc} in the models.
Finally, the suite of product understanding tasks constitutes a multi-modal problem. While current research on Large Vision Language Models (LVLMs)\citep{Qwen-VL, dai2024instructblip, zhu2023minigpt, liu2023llava, ye2023mplugowl2} can accomplish these tasks to some extent, their lack of domain knowledge in e-commerce platforms and still weak instruction following capabilities make them fall short of meeting practical requirements.

To tackle these issues, we present \textbf{\textsc{PumGPT}}, a large vision-language model expert for a series of multi-modal product understanding tasks. To be specific, we collect more than one million product data from the AliExpress platform\footnote{\url{https://www.aliexpress.com/}}, including product images, captions, categories, and lists of attributes. To filter out those attributes that cannot be inferred from product images and captions, we propose a universal hallucination detection framework utilizing multi-expert collaboration. Through the thorough hallucinated attributes filtering, we obtain about 663k data for training. Subsequently, we carefully curate five tasks that can help speed up both e-commerce platforms' and retailers' workflow. We also introduce \textbf{\textsc{PumBench}}, a benchmark covering these product understanding tasks to best evaluate the existing large vision-language models and our \textbf{\textsc{PumGPT}} in the aspect of product understanding. Extensive experiments show the \textbf{\textsc{PumGPT}} outperforms the 5 open-sourced LVLMs and GPT-4V\citep{achiam2023gpt}, the most powerful LVLM for now. And it proves the necessity of a specialized large vision language model for e-commerce.

Our contributions can be summarized as follows:
\begin{itemize}
    \item We introduce \textbf{\textsc{PumGPT}}, the first e-commerce LVLM for a series of product understanding tasks trained on a 663k high-quality product dataset with hallucination filtered.
    \item We present a universal hallucination detection framework utilizing multi-expert collaboration to detect and filter the inconsistent attributes in the dataset without any labor force.
    \item Extensive experiments demonstrate the remarkable performance of our \textbf{\textsc{PumGPT}} in \textbf{\textsc{PumBench}} over several LVLMs, including GPT-4V.
\end{itemize}

\section{Related Works}

\textbf{Vision-Language Models.} Recent advancements have shown significant success in leveraging large language models for vision-language tasks. Notable among these, Flamingo\citep{Alayrac2022FlamingoAV} employs a gated cross-attention mechanism to align vision representations with language models. Blip-2\citep{Li2023BLIP2BL} introduces a Q-Former to effectively bridge the gap between visual and textual representations. Moreover, models like Kosmos-1\citep{Huang2023LanguageIN} and PaLM-E\citep{Driess2023PaLMEAE} achieve alignment between multi-modal and text representations, creating a comprehensive interface for multi-modal input with large language models. GPT-4\citep{achiam2023gpt} has demonstrated robust visual reasoning abilities across diverse vision-linguistic tasks. Unlike end-to-end model training, some approaches coordinate multiple models to interpret and respond to multi-modal inputs, exemplified by Visual ChatGPT\citep{Wu2023VisualCT}, MM-REACT\citep{yang2023mmreact}, and HuggingGPT\citep{Shen2023HuggingGPTSA}. Increasing model sizes raise computational complexity and training data demands, prompting recent studies to explore efficient finetuning methodologies for large vision-language models\citep{zhu2023minigpt, ye2023mplugowl2, Zhang2023LLaMAAdapterEF}. Moreover, the pipeline for pretraining and instruction tuning has emerged as a new paradigm for LVLMs\citep{liu2023llava, Qwen-VL, dai2024instructblip}. However, these models often lack strict adherence to instructions, hampering their usability in large-scale e-commerce scenarios. Our \textbf{\textsc{PumGPT}} is an expert LVLM specifically trained for product understanding tasks, ideally suited for the e-commerce context.
\begin{figure*}[t]
  \includegraphics[width=\linewidth]{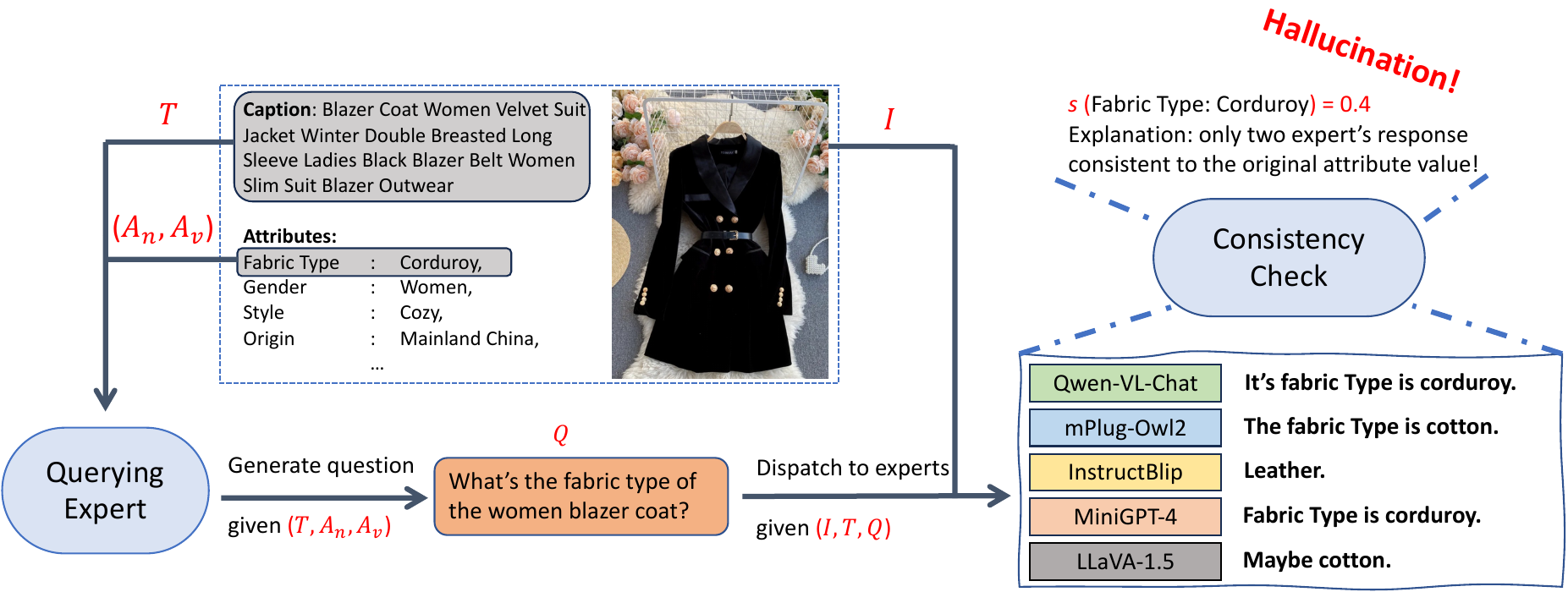} \
  \caption {The overview of our proposed hallucination detection framework.}
  \label{fig:fig3.2.1}
\end{figure*}

\noindent \textbf{Product understanding models.} Product understanding tasks encompass a variety of sub-tasks, with attribute extraction being the most extensively studied. Traditional approaches employ tagging-based models \citep{zheng2018opentag, xu2019scaling, yan2021adatag} or question-answer-based models \citep{shinzato2022simple} to extract attributes from textual product profiles. Recent research has incorporated visual information from product images to enhance attribute extraction performance \citep{lin2021pam, zhu2020multimodal, zhang2023pay}. This fusion of textual and visual data enriches the model's comprehension and extraction capabilities. Besides attribute extraction, other product understanding tasks such as product captioning \citep{atici2021generating} and product classification \citep{bonnett2016classifying} have also been explored. However, these solutions typically necessitate training separate models for each task. In contrast, our \textbf{\textsc{PumGPT}} integrates all product understanding tasks, significantly improving performance across tasks due to diverse training data and the intrinsic capabilities of \textbf{\textsc{PumGPT}}.

\section{\textsc{PumGPT}}
\begin{table}
  \centering
  \begin{tabular}{lcc}
    \hline
    \textbf{Statistical Item} & \textbf{Raw \#} & \textbf{Clean \#} \\
    \hline
    Products    & 996,350   & 663,330       \\
    Attributes     & 10,729,585   & 1,484,948       \\
    Attribute names     & 12,013   & 11,291       \\
    Attribute values     & 59,669    & 48,448       \\
    Categories     & 7,084   & 4,598       \\\hline
  \end{tabular}
  \caption{The statistical results of the raw collected data and cleaned data. We report the unique items.}
  \label{tab:tab3.1.1}
\end{table}
\subsection{Data Collection}
For sellers, an ideal process for listing products only needs to upload the product images. The system would then automatically generate attractive product titles and compile a series of product attributes for customer reference. The seller would only need to perform a final review and add any additional details if necessary. To achieve this, we gathered a total of about 1 million product entries from the AliExpress platform. Each product entry contains an image, a caption, the product category, and a set of product attributes. Each attribute consists of an attribute name and a corresponding attribute value. Table \ref{tab:tab3.1.1} demonstrates the statistical results of the raw data.

\subsection{Hallucination Filtering}
\label{sec3.2}

The initial dataset acquired from the Internet contains substantial noise stemming from multiple factors: many items lack essential product information, such as categories or attributes, making them unsuitable for training. Additionally, certain attributes might either complement product descriptions and images or conflict with other information sources due to sellers' errors. Consequently, models trained on such datasets might generate inaccuracies during inference. To mitigate this, we propose a universal hallucination detection framework aimed at filtering out noisy samples from a dataset containing approximately one million entries. This framework leverages multi-expert collaboration to identify inconsistent attributes without manual intervention.

Contemporary Large Vision Language Models (LVLMs) are pre-trained and fine-tuned on diverse datasets with varying architectures, leading to significant variability in their inference behaviours. Despite these differences, LVLMs can reach consensus on tasks requiring common knowledge or reasoning, while they generate divergent speculations when faced with ambiguous queries. This property can be exploited to detect inconsistencies within product datasets, particularly where attributes misalign with product descriptions and images. By utilizing distinct LVLMs, each with unique knowledge backgrounds, more consistent responses can be generated for accurate attribute values, whereas varied responses indicate mismatched or supplementary information or subjectively valued attributes.

As shown in Figure \ref{fig:fig3.2.1}, we selected five LVLMs as experts in hallucination detection: $\mathcal{E}$ = \{\text{Qwen-VL-Chat}\citep{Qwen-VL}, \text{MiniGPT-4}\citep{zhu2023minigpt}, \text{InstructBLIP}\citep{dai2024instructblip}, \text{mPLUG-Owl2}\citep{ye2023mplugowl2}, \text{LLaVA}\citep{liu2023llava}\}. After removing samples with missing information, a standard sample $S = (I, T, C, A_n, A_v)$ is obtained, where $I$ represents the product image, $T$ the product title, $C$ the product category, $A_n$ the attribute name, and $A_v$ the attribute value. For each attribute pair $(A_n, A_v)$, a querying expert generates questions about $A_v$. As $A_n$ is not a typed item, the Vicuna-13B\citep{vicuna2023} querying expert generates a question $Q = Vicuna(P_q, T, A_n, A_v)$ based on the attribute value type. The prompt $P_q$ for generating questions is shown in Table \ref{tab:tab.a.1.1}.

For $e_i \in \mathcal{E}$, the answer to attribute question $Q$ is formulated as $a_i = e_i(I, T, Q)$. After generating all expert answers, an additional judge checks the consistency across all answers and the original attribute value. Since experts generate answers in varied forms, they might use diverse phrases to convey the same meaning. We adopt Mistral $8\times \text{7B}$ \citep{Jiang2024MixtralOE}, a powerful large language model with a mixture of experts structure\citep{Fedus2021SwitchTS}, to evaluate the original attribute value by assigning a score $s$ from the experts as shown in Equation \ref{eq:eq3.2.1}.

\begin{equation}
  \label{eq:eq3.2.1}
  s = \sum_{e_i}^{\mathcal{E}}\frac{Mistral(e_i, A_v)}{|\mathcal{E}|}
\end{equation}

\begin{figure}[t]
  \includegraphics[width=\linewidth]{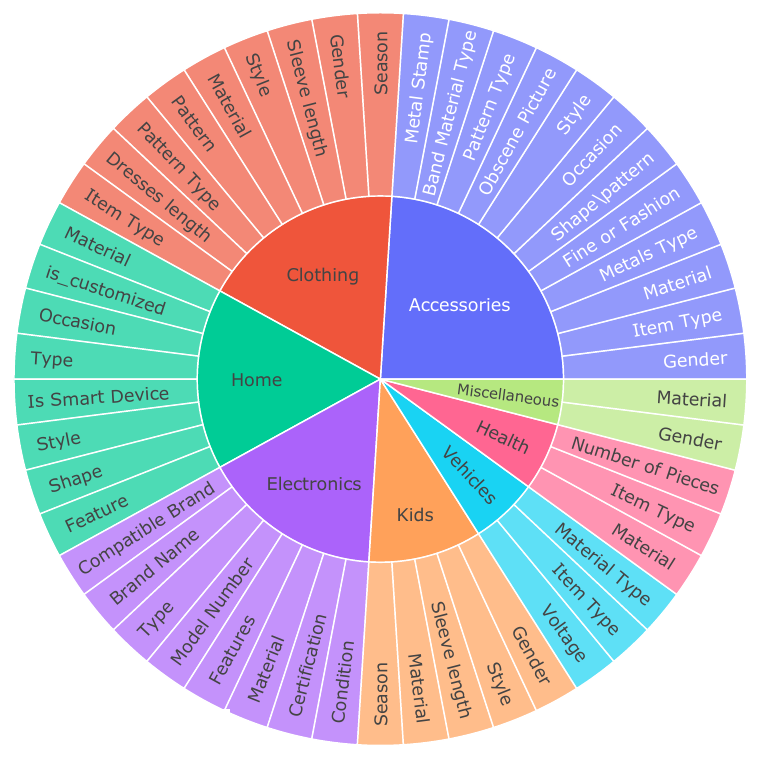}
  \caption {Most common attribute names and proportion of 8 primary categories.}
  \label{fig:fig3.2.2}
\end{figure}

Here, $Mistral(\cdot, \cdot)$ is a binary indicator function checking whether expert answers are equivalent to the original attribute value. An attribute pair is filtered as a hallucination if the score is below a threshold $\epsilon$. Practically, $\epsilon$ is set to 0.6, meaning a pair remains only when at least three experts agree with the original attribute value. Table \ref{tab:tab3.1.1} shows raw data statistics. To illustrate the training set composition, we divided over 4,000 leaf categories into eight primary ones, selecting common attribute names for each and displaying them in Figure \ref{fig:fig3.2.2}.

\subsection{Product Understanding Tasks Formulation}

\begin{table*}[h]
    \centering
    \begin{tabularx}{\textwidth}{lXX}
        \hline
        \hline
        &  \textbf{User} & \textbf{Assistant} \\
        \hline
        CG
        & What's a suitable caption for this product that would be eye-catching in the image?
        & A catchy caption is \textcolor{red}{<PROLY New Fashion Women Headband Wide Side Big Bowknot Hairband Classic Casual Turban Girls Headwear Hair Accessories>}.\\
        \hline
        CC
        & Please incorporate these characteristics: \textcolor{blue}{[<['Brand Name', 'SoundPEATS']>, <['Communication', 'True Wireless']>, <['Bluetooth Version', '5.3']>, <['Wireless Type', 'Bluetooth']>]}, into a revised version of this caption: \textcolor{blue}{<Clear Transparent Series Earbuds with Dual ENC Mic, Gaming Mode, App Support, 40H>}.
        & Reflecting the product's attributes, a revised caption would be \textcolor{red}{<SoundPEATS Clear Transparent Series Bluetooth 5.3 True Wireless Earbuds with Dual ENC Mic, Gaming Mode, App Support, 40H>}.\\
        \hline
        AI & Looking at the image along with the caption, what can you tell us about the product's attribute \textcolor{blue}{<Gender>}? If it's indistinct, please reply with 'Unknown'.& The attribute value might be \textcolor{red}{<WOMEN>}.\\
        \hline
        AC & Does the attribute \textcolor{blue}{<Material>} correctly bear the value \textcolor{blue}{<Silk>}? In case of an inaccuracy, please advise on the accurate value. & No, the product's \textcolor{red}{<Material>} attribute is \textcolor{red}{<Metal>}. \\
        \hline
        CMC & From the selection below, pick the number corresponding to the category you deem most suitable for the product: \textcolor{blue}{1: <Hair Ties>,2: <Headband>,3: <Hair Accessories Set>,4: <Hair Claw>,5: <Hair Scarf>}. & It best fits into category \textcolor{red}{<2: Headband>}.\\
        \hline
        \hline
    \end{tabularx}
    \caption{Examples of each task in the training set, where the texts in blue are the given conditions and the texts in red are the ground truth answers. Here we omit the image input.}
    \label{tab:tab3.3.1}
\end{table*}

\begin{table}
  \centering
  \begin{tabular}{lc}
    \hline
    \textbf{Tasks} & \textbf{Num of samples} \\
    \hline
    CG    & 5,000          \\
    CC     & 960        \\
    AI     & 6,031         \\
    AC     & 5,032           \\
    CMC     & 4,967         \\\hline
  \end{tabular}
  \caption{The statistics of the \textsc{PumBench}.}
  \label{tab:tab4.2.1}
\end{table}

In considering the product listing procedures within actual production environments, we have rigorously designed five tasks aimed at optimizing the efficiency of the overall production process.

\textbf{(1) Caption Generation (CG)}: The task requires the model, given an image of a product, to generate a caption that encapsulates key information about the product. \textbf{(2) Product Category Multiple-Choice Question (CMC)}: Here, the model must select the most appropriate category from a list of options, based on the product's image and caption. The options are derived from a category taxonomy tree, sourced from AliExpress, with at most nine sibling categories sampled to form the choices. \textbf{(3) Attribute Inference (AI)}: This task involves the model inferring the value of an attribute from the image and caption, based on a provided attribute name. For attributes that are challenging to determine, the model should also reject responding. To achieve this, filtered attributes are reused and their values are designated as 'Unknown'. Building upon these foundational tasks, we developed two advanced tasks. \textbf{(4) Caption Completion (CC)}: As new attributes are introduced, the model must complete the existing caption to include all necessary keywords for display. For training samples, we eliminate all keywords listed in the attributes. \textbf{(5) Attribute Correction (AC)}: The model's task is to identify and correct discrepancies between attribute values provided by the seller and other existing information about the product. In case of an error, the model should supply the correct attribute value. For practical purposes, the original value is replaced with a random one. Approximately 15 instructions and 10 response templates were designed for each task to ensure diversity. Using a conversation format akin to Qwen-VL-Chat \citep{Qwen-VL}, specific values are contained within <> to facilitate extraction in real scenarios. Table \ref{tab:tab3.3.1} offers several examples of each task, elucidating the details of these five tasks.

\section{Benchmarking on Product Understanding Tasks}
\begin{table*}[h]
\renewcommand\arraystretch{1.15}
    \centering
    \begin{tabular}{cccccccc|c}
    \hline
    \multicolumn{2}{c}{\textbf{Tasks}}& \textbf{InstBLIP} & \textbf{LLaVA} & \textbf{Mini} & \textbf{Owl2} & \textbf{Qwen-VL} & \textbf{GPT-4V} & \textbf{\textsc{PumGPT}}\\
    \hline\hline

    \multirow{3}{*}{\textbf{CG}} & Bleu$_1$ &0.094&0.069&0.086&0.087&\underline{0.153}&0.102&\textbf{0.383} \\
    &ROUGE$_L$&0.120&0.073&0.080&0.092&\underline{0.148}&0.110&\textbf{0.286} \\
    &CIDEr &0.157&0.089&0.181&0.171&\underline{0.295}&0.128&\textbf{0.987}\\

    \cline{2-9}
    \multirow{4}{*}{\textbf{CC}} & Bleu$_1$ &0.225&0.442&0.447&0.406&\underline{0.681}&0.442&\textbf{0.934} \\
    &ROUGE$_L$&0.383&0.370&0.578&0.388&\underline{0.687}&0.337&\textbf{0.937}\\
    &CIDEr &2.325&2.075&3.882&1.717&\underline{4.837}&1.281&\textbf{8.595}\\
    &Rec(\%)&6.07&32.69&18.29&40.99&47.00&\textbf{92.09}&\underline{70.63} \\

    \cline{2-9}
    \textbf{AI}&Acc(\%)&5.45&22.90&4.73&19.25&19.89&\underline{26.98}&\textbf{60.70}\\

    \cline{2-9}
    \multirow{4}{*}{\textbf{AC}} & F1(\%) &66.77&59.25&42.39&58.12&\underline{77.79}&71.38&\textbf{93.14} \\
    &Prec(\%)&50.43&54.77&65.39&60.09&69.20&\underline{81.11}&\textbf{90.34}\\
    &Rec (\%)&98.77&64.53&31.37&56.29&\underline{88.81}&63.74&\textbf{96.12}\\
    &CAcc(\%)&1.06&0.41&38.92&0.29&0.37&\underline{50.01}&\textbf{60.52} \\

    \cline{2-9}
    \textbf{CMC}&Acc(\%)&24.82&32.55&39.45&61.73&46.39&\underline{82.55}&\textbf{82.57}\\

    \hline\hline
    \end{tabular}
    \caption{The experimental results on \textsc{PumBench}, where CAcc is the accuracy of the attribute correction. We abbreviate the models for better vision effect, where InstBLIP is for InstructBLIP, Mini for MiniGPT-4, Owl2 for mPlug-Owl2, Qwen-VL for Qwen-VL-Chat. We report the results * 100\% for all the metrics except for the Bleu$_1$, ROUGE$_L$ and CIDEr. }
    \label{tab:tab5.1}
\end{table*}

\subsection{Implementation details and baselines}
\textbf{Implementation details.} We choose Qwen-VL-Chat as our base model and train with LoRA\citep{hu2022lora}, a parameter-efficient finetuning method for 3 epochs with batch size 144. The LoRA rank and alpha are 128 and 16 respectively. We employ AdamW\citep{Loshchilov2017DecoupledWD} as the optimizer. The learning rate has a linear warm-up from 1e-8 to 1e-5, followed by a cosine-decay from 1e-5 to 0. The model is trained with 8 Nvidia A100 (80G) GPUs for about 24 hours.

\noindent \textbf{Baselines.} We employ InstructBLIP\citep{dai2024instructblip}, LLaVA-1.5\citep{liu2023llava}, mPlug-Owl2\citep{ye2023mplugowl2}, MiniGPT-4\citep{zhu2023minigpt}, Qwen-VL-Chat\citep{Qwen-VL} and GPT-4V\citep{achiam2023gpt} to be the compared baselines. For both hallucination detection and evaluation on \textsc{PumBench} of all the compared methods, we set temperature and top\_p to 0.9 and 0.2 respectively. For GPT-4V, we follow its default setting. The details can be seen in Table \ref{tab:tab.a.2.1} in Appendix, and the prompts used for inference are shown in Table \ref{tab:tab.a.1.1} in Appendix.

\subsection{Datasets and metrics}

\textbf{\textsc{PumBench}.} We construct \textsc{PumBench} to evaluate the capabilities of product understanding of \textsc{PumGPT} and the existing LVLMs. We collect 1.5k items and employ 2 PhD students to clean the hallucination attributes to construct the attribute inference test set according to their commonsense. We construct other task benchmarks as we did in building the training set. The statistics of \textsc{PumBench} are shown in Table \ref{tab:tab4.2.1}.

\noindent \textbf{Metrics.}
Due to the different output formats and diverse representations of the baselines, we employ the Mistral 8$\times$7B\citep{Jiang2024MixtralOE} to serve as the answer equivalence judge to determine the accuracy of the attribute-related tasks. For CG and CC tasks, we adopt Bleu$_1$\citep{Papineni2002BleuAM}, ROUGE$_L$\citep{Lin2004ROUGEAP} and CIDEr\citep{Vedantam2014CIDErCI} metrics. Besides, we use recall as an additional metric to evaluate the CC task. We utilize accurarcy(acc), F1, precision(prec), and recall(rec) to assess the attribution correction task and only accuracy on CMC task. All reported results are the averages of three separate runs.

\section{Experimantal Results}
\subsection{Main Results on \textsc{PumBench}}

Table \ref{tab:tab5.1} elucidates the comparative performance of \textsc{PumGPT} and other methodologies on \textsc{PumBench}. Overall, \textsc{PumGPT} demonstrates superior efficacy across a variety of tasks. Specifically, in the two caption-centric tasks, \textsc{PumGPT} excels in generating captions aligned with product attributes by distilling key characteristics from images. This proficiency translates into markedly higher scores on the ROUGE$_L$ and CIDEr metrics, which evaluate recall and specific keyword utilization. In the caption completion task, aided by a base caption, \textsc{PumGPT} achieves higher performance in caption-related metrics. However, while GPT-4V successfully recalls nearly all keywords, \textsc{PumGPT} achieves a recall rate of only 70\%. This discrepancy occurs because GPT-4V formulates the completed caption from most attribute values in the reference list rather than amending the original title, resulting in GPT-4V's underperformance in caption-related metrics.

Regarding the attribute-related tasks, \textsc{PumGPT} significantly surpasses both open-source models and GPT-4V. Notably, for attribute inference task, PUMGPT exceeds the performance of GPT-4V by a margin of over thirty percentage points, highlighting the difficulties that even advanced commercial models face in intricate product understanding tasks that require specialized domain knowledge. Furthermore, due to stringent compliance regulations, GPT-4V fails to address some test samples involving prohibited topics. In the attribute correction task, \textsc{PumGPT} maintains an F1 score exceeding 90\%, while other models exhibit relatively weaker performance. Many open-source models falter in adhering to the provided instructions, thereby failing to furnish accurate values despite identifying erroneous attributes. Only MiniGPT-4 and GPT-4V can provide corrections, albeit still trailing \textsc{PumGPT}.

In the product category multiple-choice question task, \textsc{PumGPT} continued to demonstrate best-in-class performance. However, the margin was not as pronounced as in other tasks. GPT-4V's performance was comparable to \textsc{PumGPT}, suggesting that this task, which fundamentally involves reasoning rather than domain-specific knowledge, presents a fairer comparative framework. This observation implies that GPT-4V's reasoning capabilities are superior. Despite training, our model only equaled GPT-4V's performance, indicating potential areas for further enhancement in this task.

\subsection{Domain-level Results on Attribute Inference}
\label{sec5.2}
\begin{table}
  \centering
  \begin{tabular}{cccc}
    \hline
    \textbf{Tasks} & \textbf{Home} & \textbf{Electronics} & \textbf{Clothing}\\
    \hline
    InstBLIP    & 10.20 & 7.17 & 3.80          \\
    LLaVA     & 22.71 & 25.26 & 21.57        \\
    Mini     & 8.75 & 6.42 & 3.23         \\
    Owl2     & 20.00 & 18.85 & 19.24           \\
    Qwen-VL     & 14.17 & 25.01 & 17.83         \\
    GPT-4V     & \underline{29.79} & \textbf{36.04} & \underline{22.33}         \\\hline
    \textsc{PumGPT}     & \textbf{32.91} & \underline{35.49} & \textbf{78.26}         \\\hline
  \end{tabular}
  \caption{Domain-level results on attribute inference task.}
  \label{tab:tab5.2.1}
\end{table}

We divided the attribute inference task test set into three major categories: Home, Electronics, and Clothing. Both the Home and Electronics domains encompass standardized goods. For these domains, most attributes and attribute values are predefined, allowing them to be directly extracted from product titles and specifications. Consequently, a product understanding model must have thoroughly internalized this information during training to accurately infer attribute values. In contrast, Clothing items represent non-standardized goods, characterized by attributes that may be custom-defined by vendors and subject to personal interpretation. For instance, the style of a garment could be described as both commute and casual. Therefore, product understanding models must learn the distribution of vendor-specific styles during training, suggesting a higher emphasis on fitting specific distributions.

Table \ref{tab:tab5.2.1} presents the performance outcomes of each method. Overall, \textsc{PumGPT} consistently demonstrated superior performance. Within the Home domain, our results exceeded those of GPT-4V by over three percentage points, and in the Electronics domain, the margin was 0.5 percentage points. \textsc{PumGPT} outperformed the best Large Vision and Language Models (LVLMs) in standardized goods categories.

In the context of non-standardized goods, \textsc{PumGPT} showcased exceptional performance on the attribute inference task by effectively learning from product data, thus capturing the distribution of vendor-desired descriptions. Conversely, models that lacked specific training only produced results reflecting their pre-training distributions. The performance of alternative models remains inadequate for application in real-world production environments.

\subsection{Ablation on Hallucination Filtering}

\begin{figure}
  \includegraphics[width=\columnwidth]{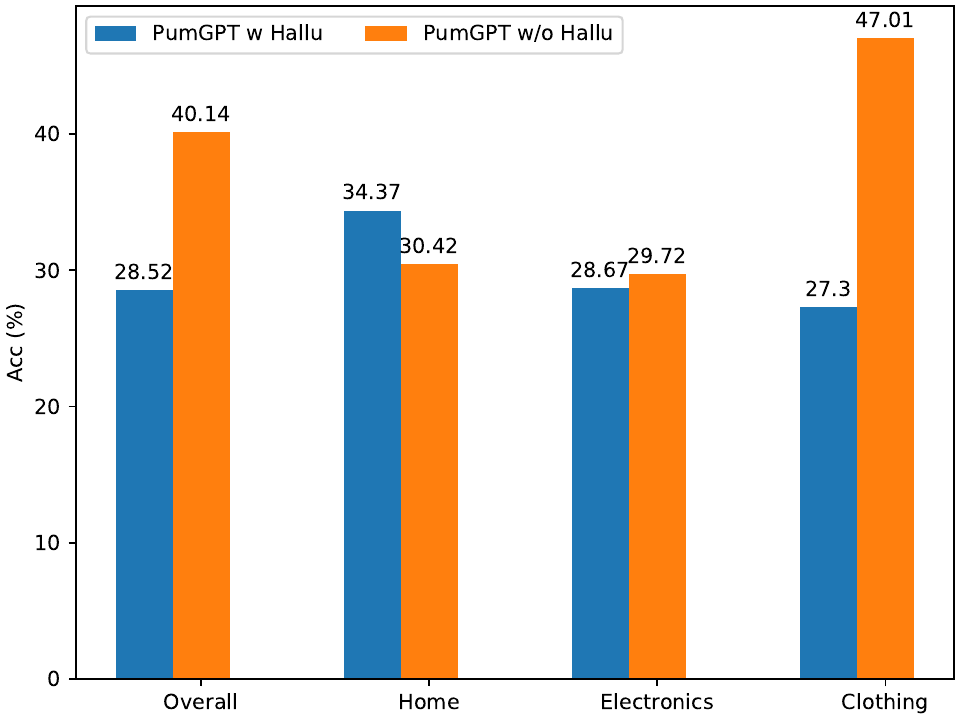}
  \caption{Ablation on hallucination filtering. Here we report the accuracy of the attribution inference task, where w Hallu means it was trained on the hallucination dataset and w/o Hallu means was trained on the hallucination-free dataset.}
  \label{fig:fig5.3.1}
\end{figure}

In Section \ref{sec3.2}, the crucial step involves filtering potentially hallucinatory attributes using our proposed multi-expert collaborative hallucination detection framework. For the task of attribute inference, \textsc{PumGPT} achieved more than double the accuracy of GPT-4V. This significant performance improvement prompted an investigation to determine if it stemmed from our handling of hallucinations and to uncover the underlying causes.

We conducted an ablation experiment on hallucination processing. A subset of 600k entries was extracted from the original dataset of 663k entries. For the dataset containing hallucinations, up to eight attributes from each product's original attribute list were randomly sampled for training. For the hallucination-free dataset, the methods outlined in Section \ref{sec3.2} were followed. The number of filtered attributes, including those designated as unknown, was strictly limited to eight. Both models underwent training for two epochs under identical training parameters.

As illustrated in Figure \ref{fig:fig5.3.1}, \textsc{PumGPT} without hallucination data (w/o Hallu) showed significant performance improvement. The accuracy was classified into three primary categories, consistent with Section \ref{sec5.2}, to elucidate distinctions. In the standardized categories, performance differences between the models were marginal. In the Home category, \textsc{PumGPT} with hallucination data (w Hallu) outperformed \textsc{PumGPT} w/o Hallu by approximately four percentage points due to learning more attributes from the dataset. However, in the Clothing category, \textsc{PumGPT} w/o Hallu exceeded the other model by nearly 20 percentage points. The Clothing category predominantly includes non-standardized clothing items, with attributes often described subjectively. Consequently, \textsc{PumGPT} trained with hallucinated data may produce excessively imaginative yet inaccurate responses. In contrast, the model trained on the hallucination-free dataset can reduce such extrapolations, resulting in more accurate responses. Therefore, the processing of hallucinations is unequivocally vital for model training.

\subsection{Evaluation on Rejection Ability}

\begin{table}
  \centering
  \begin{tabular}{ccccc}
    \hline
    \textbf{Models} & \textbf{F1} & \textbf{Prec} & \textbf{Rec} & \textbf{Acc}\\
    \hline
    InstBLIP    & 0&0&0 &89.53       \\
    LLaVA     & 17.67&\underline{20.95}&15.27  &88.30       \\
    Mini     & 0.75&4.44&0.41  &\underline{90.10}  \\
    Owl2     & 11.11&8.73&15.27 & 79.93          \\
    Qwen-VL     & 12.66&8.79&22.60  & 74.38      \\
    GPT-4V & \underline{29.69}&19.33&\textbf{64.01} & 74.47\\
    \hline
    \textsc{PumGPT} & \textbf{47.18}&\textbf{55.22}&\underline{41.12}  & \textbf{92.39}\\
    \hline
  \end{tabular}
  \caption{The evaluation on the rejection ability of all the compared methods.}
  \label{tab:tab5.4.1}
\end{table}

Large language models are acclaimed for their advanced text completion capabilities. However, they can sometimes produce incorrect information due to excessive associative reasoning. An effective model in practical applications should have the ability to refrain from responding when confronted with nonexistent or ambiguous attributes rather than providing a plausible but incorrect answer.

Consistent with our hallucination treatment within the training set, \textsc{PumGPT} defaults to the special attribute value "unknown" when queried about potentially hallucinatory attributes. As depicted in Table \ref{tab:tab5.4.1}, accuracy (acc) is measured by labeling samples that refuse to respond as 1, and those that do not as 0. If no sample is refused, the acc would be 90\%. Recall evaluates the recall rate among samples where a refusal is expected. Various models were assessed on their capacity to refuse to answer in attribute inference tasks. Open-source models like InsturctBLIP and MiniGPT-4 typically provide an actual value rather than refusing, inflating acc to around 90\%. Therefore, examining F1, precision, and recall metrics is crucial as these indicate the susceptibility of these models to hallucinations, even when instructed to refuse.

In contrast, other open-source models attempt more refusals but achieve unsatisfactory accuracy. GPT-4V demonstrates higher refusal rates due to its conservative rules, but its overall accuracy is among the lowest. While our model’s recall is lower than GPT-4V, it significantly excels in the overall F1 metric, demonstrating the effectiveness of our approach with "unknown" attributes in training sets. To enhance the model's refusal capabilities, employing preference learning algorithms such as PPO \citep{Schulman2017ProximalPO} and DPO \citep{rafailov2023direct} may be necessary.

\subsection{Case Study}
We also perform a case study in Appendix \ref{app.a.3}.

\section{Conclusion}
In this work, we introduce \textsc{PumGPT}, the pioneering Large Vision Language Model (LVLM) for e-commerce product understanding. We amassed over one million product entries and employed a multi-expert collaborative hallucination handling framework to eliminate mislabeled attributes or those not inferable from text and images. We devised five product understanding tasks aligned with actual product publishing processes, resulting in a dataset of approximately 663,000 entries to train \textsc{PumGPT}. We also developed \textsc{PumBench} to assess the performance of \textsc{PumGPT} and other LVLMs in product understanding. Experimental results reveal that \textsc{PumGPT} outperforms general-purpose LVLMs, such as GPT-4V, across all tasks. Future work will expand task variety and improve data quality to enhance model performance further.

\section*{Limitations}
Although \textsc{PumGPT} demonstrated superior performance in evaluations, it still has some limitations. (1) in the CMC task, \textsc{PumGPT}'s performance did not significantly surpass GPT-4V. Additionally, there is a considerable accuracy gap between standardized product attribute inference tasks and non-standardized product tasks. Introducing more trainable parameters or applying preference learning algorithms to specifically enhance these tasks is necessary. (2) we designed only five product understanding tasks for training, which resulted in a weaker generalization ability of the model. This limitation makes it challenging to extend to other advanced product understanding tasks, such as identifying identical products and generating product descriptions. Consequently, the model's capacity to leverage the full potential of large language models is still insufficient. To address these limitations, it is necessary to introduce a greater variety and diversity of task data. This should include not only task-specific data but also general instruction data to improve the model's generalization capability.
\bibliography{custom}

\appendix

\newpage
\section{Appendix}
\label{sec:appendix}

\subsection{Prompts}
Here we provide all the prompts used for generating attribute questions, checking equivalent attribute values, and benchmarking in table \ref{tab:tab.a.1.1}.

\subsection{Model Details}
The details of the model we compared and other generation configs are shown in Table \ref{tab:tab.a.2.1}.

\begin{table}[h]
  \centering
  \begin{tabular}{ccc}
    \hline
    \textbf{Models} & \textbf{Size} & \textbf{LLM} \\
    \hline
    InstBLIP    & 7B & Vicuna       \\
    LLaVA     & 7B& LLaMA       \\
    Mini     & 7B&LLaMA-2 \\
    Owl2     & 7B&LLaMA-2         \\
    Qwen-VL     & 7B &Qwen     \\
    GPT-4V & / & /\\
    \hline
    \textsc{PumGPT} & 7B & Qwen\\
    \hline
  \end{tabular}
  \caption{The details of model size and their base LLMs.}
  \label{tab:tab.a.2.1}
\end{table}

\subsection{Case Study}
\label{app.a.3}
We also conducted a case study. Table \ref{tab:tab.a.3.1} and Table \ref{tab:tab.a.3.2} respectively display the results of all the models for a certain attribute on non-standardized and standardized products. It can be observed that most models are unable to infer results for the non-standardized product. These models either fail to generate the results or mistakenly output the entire product title while intending to express prominent text on the clothes, leading to errors. However, \textsc{PumGPT} effectively avoided this issue and accurately inferred the correct attribute values.

For the standardized product, the attribute "Model Number" is challenging to determine. Consequently, almost all models performed poorly. Other models directly refused to answer, while \textsc{PumGPT} attempted to extract a reasonable model number from the title. Despite this effort, it similarly repeated the entire title, as observed in the previous case. This indicates that \textsc{PumGPT} still has deficiencies in extracting complex attributes. Addressing this issue may require more difficult samples for training.

\begin{table*}[h]
    \centering
    \begin{tabularx}{\textwidth}{lX}
        \hline
        \hline
        &  \textbf{Prompt} \\
        \hline
        Question Gen & Given the title of a product and a pair of attribute name and value of the product, generate a possible question about the attribute name from which the attribute value can be inferred. The question generated should not contain the attribute value and use a brief name(e.g. just a noun) to refer the product itself.

        Example:

        Product name: 4MP 1080P IP Outdoor WiFi Security Camera for Home Surveillance, Waterproof Bullet Cam, HD WiFi Video. Attribute name: Supported Mobile Systems. Attribute value: Android. Question: What is the supported mobile systems of the camera?

        Product name: <>. Attribute name: <>. Attribute value: <>. Question: \\
        \hline
        Answer Check& Given a certain attribute of a product, you're required to judge whether a candidate attribute value is completely equivalent to the reference attribute value without any ambiguity (consistent keywords and the same number of keywords). Simply respond with "yes" (indicating the two values are equivalent) or "no" (indicating they're not).

        Attribute name: <>. Reference attribute value: <>. Candidate attribute value: <>.

        Judgement: \\
        \hline
        CG
        & Generate a caption that encapsulates the essence of the product in the image.
        \\
        \hline
        CC
        & Enhance this existing title to make it more appealing for the product shown in the image with these features given: [<{}>]. The initial caption is: <{}>. Just answer the enhanced caption based on the initial caption with necessary attributes.
        \\
        \hline
        AI & The caption of the product in the image is {}. Please clarify the attribute {} of the product. Just respond with a simple phrase and respond unknown if you're not sure.\\
        \hline
        AC & Upon reviewing the product titled <{}> in the attached image, is the <{}> attribute correctly <{}>? Just respond yes or no. If this is incorrect, kindly provide the accurate value. \\
        \hline
        CMC & The caption of the product in the image is {}. Choose the most fitting category for the product: [<{}>]. Just answer the option number that you believe correct.\\
        \hline
        \hline
    \end{tabularx}
    \caption{The prompt used for generating attribute questions, checking equivalent attribute values, and benchmarking.}
    \label{tab:tab.a.1.1}
\end{table*}

\begin{table*}[h]
     \begin{center}
     \begin{tabularx}{\textwidth}{XXX}
     \hline
      Product & Models & Answers \\
    \hline
    \multirow{8}{*}{\raisebox{-\totalheight}{\includegraphics[width=0.3\textwidth, height=55mm]{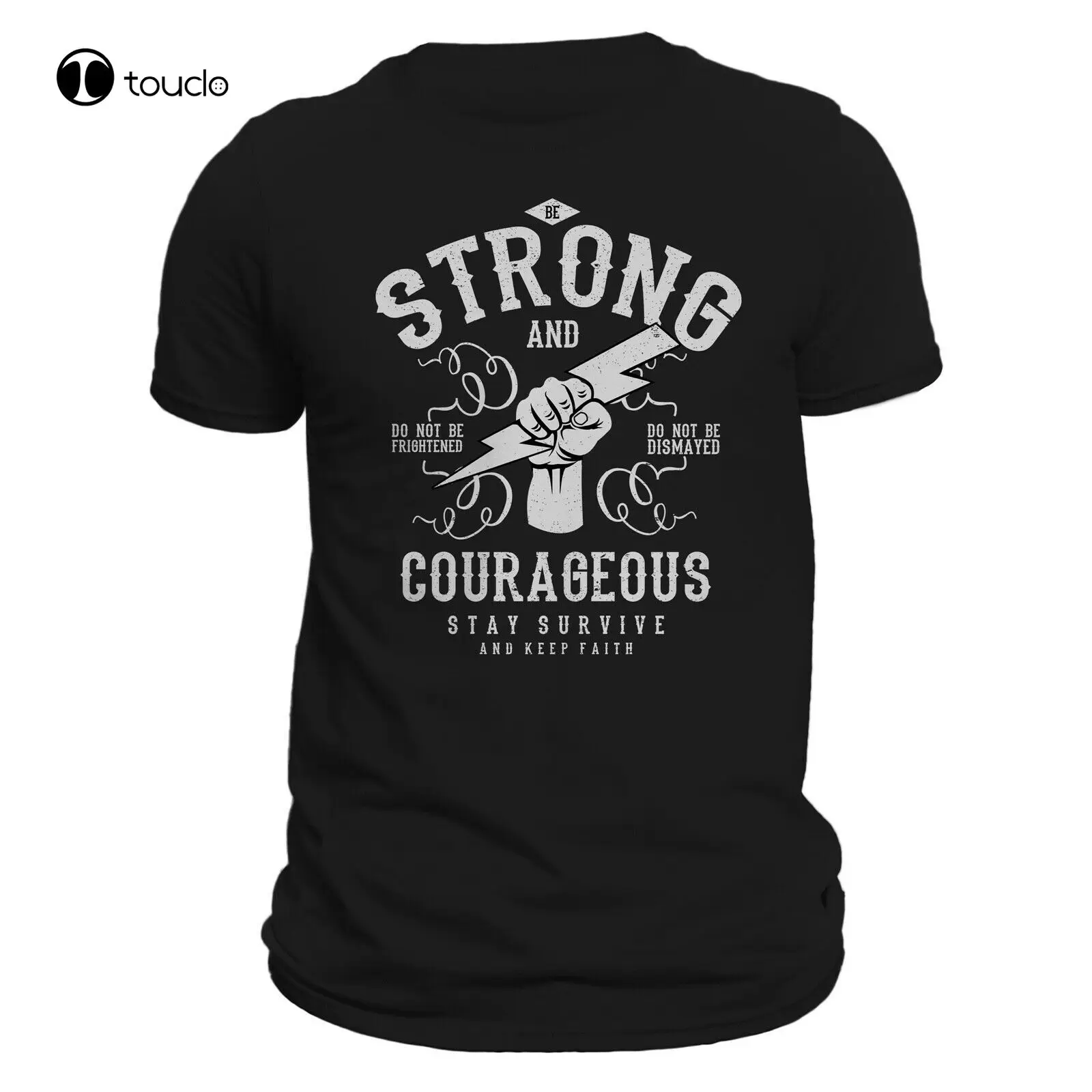}}}
      &
      InstructBLIP
      &
    What you're looking for Be Strong And Courageous Pandemicc Political Insurrection Men'S T-Shirt Custom Aldult Teen Unisex Digital Printing Xs-5Xl.\\ \cline{2-3}

      &
      LLaVA-1.5
      & Unisex - Youth.\\ \cline{2-3}

      &
      MiniGPT-4
      & The image shows a black t-shirt with the words "Be Strong And Courageous".\\ \cline{2-3}

      &
      mPlug-Owl2
      & The product is a black t-shirt featuring the phrase "Be Strong and Courageous" and a picture of a fist. It is sold by the brand toulo, available in sizes XS to 5XL. \\ \cline{2-3}

      Caption: Be Strong And Courageous Pandemicc Political Insurrection Men'S T-Shirt Custom Aldult Teen Unisex Digital Printing Xs-5Xl.
      &
      Qwen-VL-Chat
      & The pattern of the product is a simple phrase "Be Strong And Courageous Pandemicc Political Insurrection Men'S T-Shirt Custom Aldult Teen Unisex Digital Printing Xs-5Xl".\\ \cline{2-3}

    Attribute name: Pattern
      &
      GPT-4V
      & Unknown.\\ \cline{2-3}

        &
      \textsc{PumGPT}
      & The attribute value of the product may be <Print>.\\ \cline{2-3}

      &
      \textbf{GT}
      & Print.\\ \hline

      \end{tabularx}
      \caption{A case on a non-standardized product, where GT is the reference attribute value}
      \label{tab:tab.a.3.1}
      \end{center}
\end{table*}

\begin{table*}[h]
     \begin{center}
     \begin{tabularx}{\textwidth}{XXX}
     \hline
      Product & Models & Answers \\
    \hline
    \multirow{4}{*}{\raisebox{-\totalheight}{\includegraphics[width=0.3\textwidth, height=46mm]{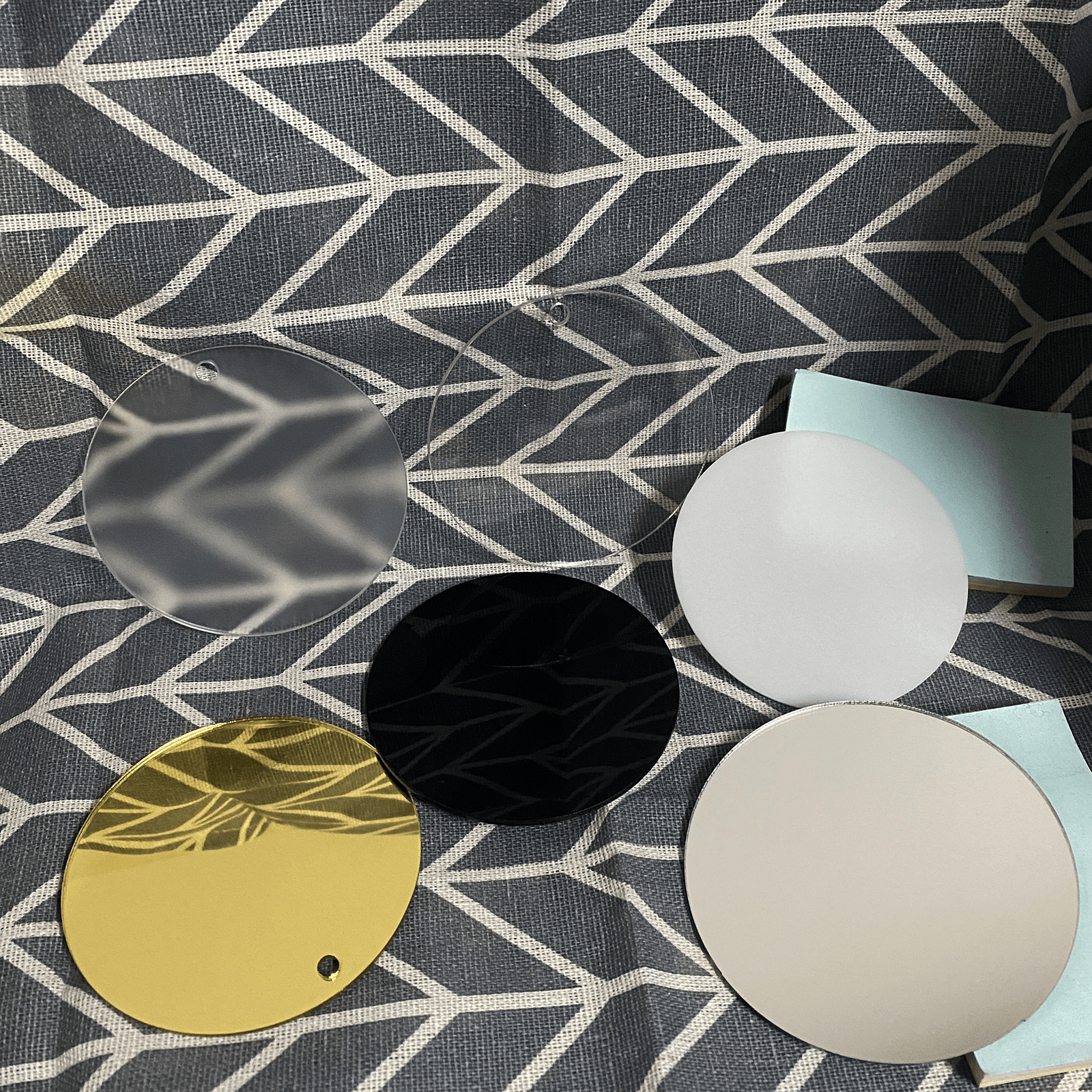}}}
      &
      InstructBLIP
      & No answer.\\ \cline{2-3}

      &
      LLaVA-1.5
      & Unknown.\\ \cline{2-3}

      &
      MiniGPT-4
      & Thank you for providing the image. The product in the image is an acrylic circle sheet disc.\\ \cline{2-3}

      &
      mPlug-Owl2
      & The model number of the product is not visible in the image. \\ \cline{2-3}

      &
      Qwen-VL-Chat
      & The model number of the product is not mentioned in the not provided.\\ \cline{2-3}

      &
      GPT-4V
      & Unknown.\\ \cline{2-3}

        Caption: 2/3/4/5/6" Acrylic Circle Sheet Disc Round Board for Wedding Place Card Gift Tag Ornament DIY Craft Art Project Painting Carve.
        &
      \textsc{PumGPT}
      & The attribute value of the product may be <2/3/4/5/6" Acrylic Circle Sheet Disc Round Board for Wedding Place Card Gift Tag Ornament DIY Craft Art Project Painting Carve>. \\ \cline{2-3}

    Attribute name: Model Number
      &
      \textbf{GT}
      & Acrylic Circle Sheet Disc Round Board.\\ \hline

      \end{tabularx}
      \caption{A case on a standardized product.}
      \label{tab:tab.a.3.2}
      \end{center}
\end{table*}

\end{document}